\documentclass[conference]{IEEEtran}
\IEEEoverridecommandlockouts
\usepackage{cite}
\usepackage{amsmath,amssymb,amsfonts}
\usepackage{amsmath}
\usepackage{algorithmic}
\usepackage{graphicx}
\usepackage{textcomp}
\usepackage{xcolor}
\usepackage{booktabs}
\usepackage{algorithm}
\usepackage{algorithmic}   % »òÕß algorithmicx
\usepackage{subcaption}   % Ìá¹© subfigure »·¾³

\usepackage{bm}           % ´ÖÌåÊýÑ§·ûºÅ£¨½â¾ö \bm ÎÊÌâ£©
\usepackage{diagbox}      % ±í¸ñÐ±Ïß£¨ÄúÖ®Ç°ÒÑÓÃµ«Ã»¼ÓÔØ£©
\usepackage{float}        % Í¼Æ¬¸¡¶¯¿ØÖÆ
\usepackage{caption}      % Í¼±í±êÌâ
\usepackage{setspace}     % ÐÐ¾à¿ØÖÆ
\usepackage{enumitem}     % ÁÐ±í¿ØÖÆ£¨Ìæ´ú enumerate£©
\usepackage{hyperref}     % ³¬Á´½Ó£¨¿ÉÑ¡£©
\newtheorem{proposition}{Proposition}

\DeclareMathOperator{\diag}{diag}
\DeclareMathOperator{\argmin}{argmin}

\DeclareMathOperator{\tr}{tr}

\def\BibTeX{{\rm B\kern-.05em{\sc i\kern-.025em b}\kern-.08em
    T\kern-.1667em\lower.7ex\hbox{E}\kern-.125emX}}
\begin{document}

\title{DK-GBMKKM: Dynamic Kernel-Space Granular-Ball Multiple Kernel $k$-Means Clustering\\
\thanks{
%	This work was supported in part by the National Natural Science Foundation of China under Grant 62221005, Grant 62450043, Grant 62222601, and Grant 62176033; in part by Chongqing University of Posts and Telecommunications Ph.D. Innovative Talent Project under Grant BYJS202403; and in part by Chongqing Natural Science Foundation Innovation and Development Joint Fund under Grant CSTB2025NSCQ-LZX01411.
X. Lian, Y. Zhang, S. Xia, S. Zhong \& X. Xiang are with the Chongqing Key Laboratory of Computational Intelligence, Key Laboratory of Cyberspace Big Data Intelligent Security, Ministry of Education, Sichuan-Chongqing Co-construction Key Laboratory of Digital Economy Intelligence
and Key Laboratory of Big Data Intelligent Computing, Chongqing University of Posts and Telecommunications, 400065, Chongqing, China. }
}

\author{\IEEEauthorblockN{1\textsuperscript{st} Xiaoyu Lian}
\IEEEauthorblockA{
%	\textit{Chongqing Key Laboratory of Computational Intelligence}\\ 
%	\textit{Key Laboratory of Cyberspace Big Data Intelligent Security, Ministry of Education}\\ \textit{Sichuan-Chongqing Co-construction Key Laboratory of Digital Economy Intelligence}\\
%	\textit{Key Laboratory of Big Data Intelligent Computing} \\
\textit{Chongqing University of Posts and Telecommunications}\\
Chongqing, China, lianxiaoyu724@qq.com}
\and
\IEEEauthorblockN{2\textsuperscript{nd} Yuchao Zhang}
\IEEEauthorblockA{
%	\textit{dept. name of organization (of Aff.)} \\
\textit{Chongqing University of Posts and Telecommunications}\\
Chongqing Open University\\
Chongqing, China, weiyanshiai@qq.com}
\and
\IEEEauthorblockN{3\textsuperscript{rd} Shuyin Xia*}
\IEEEauthorblockA{
%	\textit{Chongqing Key Laboratory of Computational Intelligence}\\ 
%	\textit{Key Laboratory of Cyberspace Big Data Intelligent Security, Ministry of Education}\\ \textit{Sichuan-Chongqing Co-construction Key Laboratory of Digital Economy Intelligence}\\
%	\textit{Key Laboratory of Big Data Intelligent Computing} \\
	\textit{Chongqing University of Posts and Telecommunications}\\
	Chongqing, China, xiasy@cqupt.edu.cn}
\and
\IEEEauthorblockN{4\textsuperscript{th} Siqi Zhong}
\IEEEauthorblockA{
\textit{Chongqing University of Posts and Telecommunications}\\
Chongqing, China, 3572288058@qq.com}
\and
\IEEEauthorblockN{5\textsuperscript{th} Zhaoxu Xiang}
\IEEEauthorblockA{
\textit{Chongqing University of Posts and Telecommunications}\\
Chongqing, China, 2130453184@qq.com}
%\and
}

\maketitle

\begin{abstract}
Multiple kernel $k$-means integrates complementary nonlinear similarities by learning a combination of base kernels. Its pointwise optimization, however, is sensitive to noisy and boundary samples and repeatedly operates on sample-scale kernel matrices. Granular-ball representations organize local sample groups into mesoscopic units, but granular balls generated once in the input space may be inconsistent with the fused-kernel geometry that evolves during multiple kernel learning. We propose dynamic kernel-space granular-ball multiple kernel $k$-means (DK-GBMKKM). The method generates granular balls in the current fused kernel space and alternates kernel-weight learning with granular-ball membership updates, allowing the representation to adapt to changes in the fused-kernel geometry. A sample-size-weighted granular-ball kernel is further constructed to preserve the contributions of balls of different sizes, and its positive semidefiniteness and related equivalence properties are established. Experiments on 12 public datasets demonstrate the strong overall clustering performance of DK-GBMKKM. The code has been open-sourced for reproducibility: https://github.com/lianxiaoyu724/DK-GBMKKM.
\end{abstract}


\section{Introduction}
Classical $k$-means clustering is widely used in image analysis, text mining, bioinformatics, and multimedia processing because of its simplicity, computational efficiency, and ease of implementation~\cite{yang2024federated,zhang2025structured,heidari2024novel}.
% ,zhang2024speeding,zhang2025structured
To overcome the limitations of Euclidean distance in the input space, spectral clustering constructs a sample-similarity graph, reformulates clustering as graph partitioning, and derives a low-dimensional embedding from eigenvectors of the graph Laplacian~\cite{liu2018spectral,ding2024survey,nie2024novel}. Kernel $k$-means instead maps samples implicitly into a reproducing kernel Hilbert space (RKHS) and clusters in that space, improving the representation of nonlinear structures~\cite{dhillon2004kernel,zhou2022memory}.
% ,zhang2002large

Although kernel clustering can capture nonlinear structures, the expressive power of a single kernel is limited. Multiple kernel clustering combines several base kernels to learn task-adaptive similarities~\cite{liu2022simplemkkm,du2015robust}. Multiple kernel $k$-means (MKKM) is a representative framework that jointly optimizes kernel weights and cluster assignments~\cite{liu2019multiple}, with subsequent extensions improving robustness, structural modeling, and scalability~\cite{wang2024multiple,liang2024consistency}. Nevertheless, most existing methods remain sample-centric, making them sensitive to noisy, boundary, and outlying samples and computationally expensive for large kernel matrices. A stable and efficient mesoscopic representation is therefore highly desirable.

Granular-ball computing (GBC) has recently emerged as an adaptive multigranularity representation. Rather than learning directly from individual samples, GBC approximates an arbitrary data distribution with granular balls characterized by centers, radii, and sample coverage. It thus replaces many samples with fewer and more stable mesoscopic units~\cite{xia2019granular,xia2022efficient}. GBC offers inherent advantages in efficiency, robustness, and interpretability and has been integrated with classifiers, rough sets, fuzzy sets, and graph learning to build stable multigranularity learning frameworks~\cite{xia2024gbsvm,xia2023gbrs,lian2026gbfrs,xia2024granular,dai2025adaptive}. It has also been used in clustering to reduce computational complexity and improve robustness to noise~\cite{xie2024mgnr,xia2025gbct,cheng2026fast,jia2025generation,chen2025gbsk,su2025multi,cheng2024gb,xie2024efficient,xie2024w}. Introducing GBC into multiple kernel clustering is therefore a natural direction. Granular-ball-induced multiple kernel k-means (GB-MKKM)~\cite{xia2025GBMKKM} first embedded granular balls into MKKM. By constructing balls in the input space and compressing the sample set, it improved efficiency and robustness and demonstrated the feasibility of mesoscopic units for multiple kernel clustering.

GB-MKKM and related methods nevertheless construct granular balls in the input space. This design implicitly assumes that samples close in the input space remain close in the high-dimensional feature spaces induced by the kernels. The assumption is reasonable only for approximately order-preserving mappings, such as a linear kernel. Under commonly used nonlinear kernels, including Gaussian and polynomial kernels, neighborhood relations can change substantially after mapping. Input-space ball boundaries may then fail to reflect the actual local density and cluster structure in kernel space. In addition, a fixed granular-ball partition cannot adapt as the kernel weights and fused-kernel geometry evolve. To address these limitations, we propose dynamic kernel-space granular-ball multiple kernel $k$-means (DK-GBMKKM), which aligns granular-ball construction with the space in which multiple kernel clustering is optimized. The main contributions are as follows:
\begin{itemize}
	\item We propose a fused-kernel-driven dynamic granular-ball generation mechanism, where granular balls are constructed and updated directly in the current fused kernel space while keeping the ball number fixed.
	\item We develop an alternating optimization framework that jointly updates the granular-ball structure, spectral representation, and kernel weights in the compressed ball space.
	\item Extensive experiments on 12 public datasets show that DK-GBMKKM outperforms seven representative methods in terms of average performance and overall ranking across four clustering metrics.
\end{itemize}

\section{Dynamic Kernel-Space Granular-Ball Multiple Kernel $k$-Means}
\label{sec:framework}

Rather than applying multiple kernel clustering directly to sample-level kernel matrices, we construct multiple granular-ball kernels and perform clustering on mesoscopic units that encode local structure.

\subsection{Motivation}
\begin{figure*}[t]
	\centering
	\includegraphics[width=0.8\textwidth]{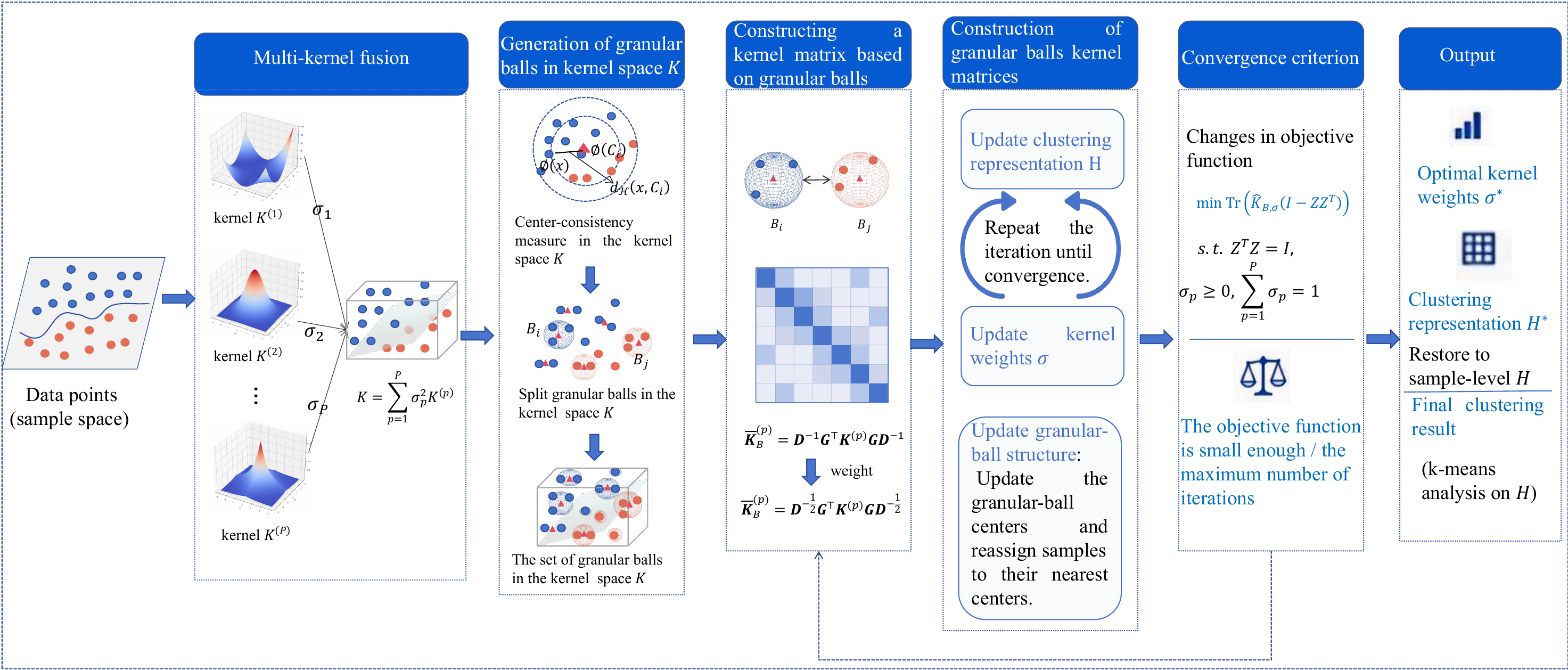}
	\caption{Framework of DK-GBMKKM.}
	\label{GBMKKM}
\end{figure*}

Existing granular-ball multiple kernel clustering methods typically construct granular balls in the input space and then build ball-level kernel representations. This causes a space mismatch, as the ball structure follows input-space geometry while clustering is optimized in the fused kernel space. Moreover, fixed granular-ball partitions cannot adapt to changing kernel weights and may become inconsistent with the evolving fused-kernel geometry. DK-GBMKKM addresses both issues, as illustrated in Fig.~\ref{GBMKKM}. It first builds a fused kernel matrix $K$ from the current kernel weights and generates granular balls in the induced kernel space. Ball-level kernel matrices are then computed from within- and between-ball kernel relations, followed by spectral embedding and kernel-weight learning at the ball level. After the weights are updated, the fused kernel is reconstructed and the granular-ball memberships are adjusted. The representation and the multiple kernel model therefore evolve together.

\subsection{Problem Formulation and Base Model}
\label{sec:preliminaries}

Let $\mathcal X=\{x_i\}_{i=1}^{n}$ be a dataset with $c$ target clusters. Given $P$ positive semidefinite base kernel matrices $\{K^{(p)}\}_{p=1}^{P}$, each of which has been symmetrized, centered, and diagonal-normalized, let $K^{(p)}\in\mathbb R^{n\times n}$. The kernel-weight vector $\sigma=[\sigma_1,\ldots,\sigma_P]^\top$ satisfies
\begin{equation}
	\left\{
	\sigma\mid
	\sigma_p\geq0,\ 
	\sum_{p=1}^{P}\sigma_p=1\right\}.
	\label{eq:simplex}
\end{equation}
We use the squared-weight kernel combination
\begin{equation}
	K_{\sigma}
	=
	\sum_{p=1}^{P}\sigma_p^2K^{(p)}.
	\label{eq:sample_fusion}
\end{equation}

\textbf{Kernel-space granular ball:}
In the RKHS $\mathcal H$ induced by the current fused kernel $K_{\sigma}$, the $\ell$th granular ball is denoted by $B_\ell=(\mathcal{I}_\ell,C_\ell,R_\ell,\text{CCM}_\ell)$. Here, $\mathcal{I}_\ell$ is the set of samples covered by the ball, $n_\ell=|\mathcal{I}_\ell|$ is its size, $C_\ell$ and $R_\ell$ are its center and radius in kernel space, and $\text{CCM}_\ell$ is its kernel-space center-consistency measure. The corresponding derivation is provided in Appendix~\ref{app:fusion_ball}.

The kernel-space center is the mean of the mapped samples in the ball:
\begin{equation}
	C_\ell
	=
	\frac{1}{n_\ell}
	\sum_{x_i\in\mathcal{I}_\ell}
	\phi(x_i),
	\label{eq:kernel_space_ball_center}
\end{equation}
where $\phi(x_i)$ is the implicit mapping of $x_i$ into the RKHS induced by $K_{\sigma}$. For any sample $x$, its squared distance to $C_\ell$ is
\begin{equation}
	\begin{aligned}
		&d_{\mathcal{H}}^2(x,C_\ell)
		=
		\left\|
		\phi(x)-C_\ell
		\right\|_{\mathcal{H}}^2\\
		&=
		K_{\sigma}(x,x)
		-
		\frac{2}{n_\ell}
		\sum_{x_i\in\mathcal{I}_\ell}
		K_{\sigma}(x,x_i)+
		\frac{1}{n_\ell^2}
		\sum_{x_i\in\mathcal{I}_\ell}
		\sum_{x_j\in\mathcal{I}_\ell}
		K_{\sigma}(x_i,x_j).
	\end{aligned}
	\label{eq:kernel_space_sample_to_center}
\end{equation}
Thus, the distance is obtained entirely from $K_{\sigma}$ without explicitly computing a center vector. The maximum and mean radii of $B_\ell$ are
\begin{equation}
	R_\ell
	=
	\max_{x_i\in\mathcal{I}_\ell}
	d_{\mathcal{H}}(x_i,C_\ell),
	\qquad
	\bar R_\ell
	=
	\frac{1}{n_\ell}
	\sum_{x_i\in\mathcal{I}_\ell}
	d_{\mathcal{H}}(x_i,C_\ell).
	\label{eq:kernel_space_radii}
\end{equation}

Because label purity is unavailable in unsupervised clustering, we adapt the center-consistency measure~\cite{xia2025gbct} by replacing all sample distances with kernel-space distances. Let $\chi_\ell=\{x_i\in\mathcal{I}_\ell\mid d_{\mathcal{H}}(x_i,C_\ell)\leq \bar R_\ell\}$ denote the samples within the mean radius. The dimension of a kernel space is generally unavailable explicitly, so we use radius-normalized densities and thereby avoid a direct dependence on dimensionality. The consistency of a singleton or zero-radius ball is set to 1. The densities within the mean and maximum radii are
\begin{equation}
	\rho_\ell^{\text{ave}}
	=
	\frac{|\chi_\ell|}{\bar R_\ell},
	\qquad
	\rho_\ell^{\max}
	=
	\frac{n_\ell}{R_\ell}.
	\label{eq:kernel_space_densities}
\end{equation}
These ratios quantify compactness by relating the number of covered samples to the corresponding radius. The kernel-space center-consistency measure is then
\begin{equation}
	\text{CCM}_\ell
	=
	\frac{
		\min(\rho_\ell^{\text{ave}},\,\rho_\ell^{\max})
	}{
		\max(\rho_\ell^{\text{ave}},\,\rho_\ell^{\max})
	}
	\in(0,1].
	\label{eq:kernel_space_con}
\end{equation}
A value close to 1 indicates a uniform and stable kernel-space distribution, whereas a smaller value suggests that the ball should be refined. The granular balls are generated in the current fused kernel space by combining this measure with kernel 2-means in the GBC procedure~\cite{xia2025gbct}.

\textbf{Granular-ball kernel construction:}
%The current fused kernel $K_{\sigma}$ determines a partition shared by all base kernels. Because the fused kernel is a weighted combination of the base kernels, a ball center in the fused kernel space has a consistent decomposition into its counterparts in the base-kernel spaces; the derivation is given in Appendix~\ref{app:fused_ball_center}. Hence, only one granular-ball partition is generated in the fused kernel space. Once fixed, the same index sets $\{\mathcal{I}_b\}_{b=1}^{M}$ are used for every base kernel.
The current fused kernel $K_{\sigma}$ induces a single granular-ball partition shared by all base kernels. Since $K_{\sigma}$ is a weighted combination of the base kernels, each fused-space ball center admits a consistent decomposition in the corresponding base-kernel spaces; see Appendix~\ref{app:fused_ball_center}. Therefore, only one partition is generated, and the same index sets $\{\mathcal{I}_b\}_{b=1}^{M}$ are used for all base kernels.

For two balls $B_a$ and $B_b$, their similarity in the $p$th base-kernel space is defined as the inner product between their centers:
\begin{equation} 
	\begin{aligned} 
		\overline{K}_B^{(p)}(a,b)
		&= \left\langle C_a^{(p)},C_b^{(p)} \right\rangle = \frac{1}{n_an_b}
		\sum_{x_i\in\mathcal{I}_a}
		\sum_{x_j\in\mathcal{I}_b}
		K^{(p)}(x_i,x_j).
	\end{aligned} 
	\label{eq:ball_center_kernel} 
\end{equation}
This kernel-space inner product is computed solely from a sample-level kernel matrix; the base-space centers need not be formed explicitly.

Define the sample-to-ball indicator matrix $G$ by
\begin{equation}
	G_{ib}
	=
	\begin{cases}
		1,&x_i\in B_b,\\
		0,&\text{otherwise},
	\end{cases}
	\qquad
	\sum_{b=1}^{M}G_{ib}=1,
	\quad i=1,\ldots,n.
\end{equation}
Thus, $G\in\{0,1\}^{n\times M}$. Define the ball-size matrix as
\begin{equation}
	D
	=
	G^\top G
	=
	\diag(n_1,\ldots,n_M).
\end{equation}
By Eq.~\eqref{eq:ball_center_kernel}, the average granular-ball kernel for base kernel $p$ is
\begin{equation}
	\overline{K}_B^{(p)}
	=
	D^{-1}
	G^\top
	K^{(p)}
	G
	D^{-1},
	\label{eq:average_ball_kernel}
\end{equation}
whose the $(a,b)$ entry is $\langle C_a^{(p)},C_b^{(p)}\rangle$.

%Eq.~\eqref{eq:average_ball_kernel} ignores ball sizes. We therefore scale the average kernel symmetrically by the square roots of the ball sizes, while omitting the common factor $1/n$ since $n$ is constant.
Eq.~\eqref{eq:average_ball_kernel} measures kernel-space similarity between ball centers but ignores ball sizes, which may underweight larger balls. We therefore scale the average kernel symmetrically by the square roots of the ball sizes, omitting the common factor $1/n$ since the total sample size is fixed.
% Eq.~\eqref{eq:average_ball_kernel} captures the kernel-space similarity between ball centers but ignores the number of samples represented by each ball. Treating all balls as equally important can therefore underrepresent larger balls. We symmetrically scale the average kernel by the square roots of the ball sizes. The common factor $1/n$ can be omitted because the total sample size is constant across balls:
\begin{equation}
	\begin{aligned}
		\widehat{K}_B^{(p)}
		&=
		D^{\frac{1}{2}}
		\overline{K}_B^{(p)}
		D^{\frac{1}{2}}=
		D^{-\frac{1}{2}}
		G^\top
		K^{(p)}
		G
		D^{-\frac{1}{2}}
		=
		Q^\top K^{(p)} Q.
	\end{aligned}
	\label{eq:weighted_ball_kernel}
\end{equation}
Its $(a,b)$th entry satisfies
\begin{equation}
	\begin{aligned}
		&\widehat{K}_B^{(p)}(a,b)
		=
		\sqrt{n_an_b}
		\left\langle C_a^{(p)},C_b^{(p)}\right\rangle,\\
		&Q=G D^{-\frac{1}{2}},
		\qquad
		Q^\top Q
		=
		I_M.
	\end{aligned}
	\label{eq:weighted_ball_kernel_element}
\end{equation}
The weighted kernel retains the similarity between ball centers while encoding the number of samples represented by each ball.

Consistent with the sample-level kernel combination, the fused granular-ball kernel is
\begin{equation}
	\widehat{K}_{B,\sigma}
	=
	\sum_{p=1}^{P}
	\sigma_p^2
	\widehat{K}_B^{(p)}.
	\label{eq:fused_ball_kernel}
\end{equation}
The original $n\times n$ multiple kernel representation is thereby converted into an $M\times M$ representation.

\begin{proposition}
If $K^{(p)}\succeq 0$, then $\widehat{K}_B^{(p)}\succeq 0$. Moreover, for any $\sigma$ satisfying Eq.~(\ref{eq:simplex}), the fused granular-ball kernel $\widehat{K}_{B,\sigma}$ is positive semidefinite.
\end{proposition}

This property ensures that the constructed matrices are valid kernels and can be used directly for kernel clustering and multiple kernel learning. The proof is given in Appendix~\ref{app:prof1}.

\begin{proposition}
If $ Z^\top Z= I_c$ and $H= Q $, then $H^\top H= I_c$,
and for any base kernel $K^{(p)}$,
\begin{equation}\label{eq:alignment_equivalence}
\tr(H^\top K^{(p)} H) = \tr(Z^\top \widehat{K}_B^{(p)} Z).
\end{equation}
\end{proposition}

As shown in Appendix~\ref{app:prof2}, ball-level spectral optimization is the projection of sample-level spectral optimization onto the subspace induced by the granular-ball partition. Samples in the same ball share an embedding vector scaled by $n_b^{-1/2}$. Dynamic ball updates therefore modify the feasible subspace of the sample embedding and, in turn, the optimized ball-level spectral representation; they do not introduce an independent free parameter outside the multiple kernel clustering model.

\subsection{Objective and Optimization of DK-GBMKKM}

Let $ Z\in\mathbb R^{M\times c}$ be the ball-level spectral embedding with $ Z^\top Z= I_c$. For the current granular-ball partition, DK-GBMKKM solves
\begin{equation}
	\begin{aligned}
		\min_{ Z,\sigma}\quad
		&J_B( Z,\sigma)
		=\sum_{p=1}^P\sigma^2_pL_p,\\
		\mathrm{s.t.}\quad
		& Z^\top Z= I_c,\qquad
		\sigma_p\geq 0,\qquad
		\sum_{p=1}^P\sigma_p=1,
	\end{aligned}
	\label{eq:ball_mkkm_objective}
\end{equation}
where $L_p=\tr(\widehat{K}^{(p)}_{B})-\tr( Z^\top\widehat{K}^{(p)}_{B} Z)$. Let $Z^{(t)}$, $\sigma^{(t)}$, $B^{(t)}$ and $J^{(t)}$ be the results after the synchronized update at iteration $t$. For fixed kernel weights $\sigma^{(t)}$, Eq.~\eqref{eq:ball_mkkm_objective} is equivalent, with respect to $Z^{(t)}$, to
\begin{equation}
	\max_{ {Z^{(t)}}^\top Z^{(t)}=I_c}
	\tr( {Z^{(t)}}^\top\widehat{K}_{B^{(t)},\sigma^{(t)}} Z^{(t)}).
\end{equation}
Accordingly, $Z^{(t)}$ consists of the orthonormal eigenvectors associated with the $c$ largest eigenvalues of $\widehat{K}_{B^{(t)},\sigma^{(t)}}$:
\begin{equation}
	 Z^{(t)}
	=
	\operatorname{Spec}(\widehat{K}_{B^{(t)},\sigma^{(t)}},c).
\end{equation}

For fixed $Z^{(t)}$, the weight subproblem is a convex quadratic program over the probability simplex. To avoid numerical instability when a residual approaches zero, define $\widetilde L_p=\max(L_p,e^{-12})$. The kernel weights are updated as
\begin{equation}
	\sigma_p^{(t+1)}
	=
	\frac{1/\widetilde L_p}
	{\sum_{q=1}^{P}1/\widetilde L_q}.
	\label{eq:weight_update}
\end{equation}
The derivation is provided in Appendix~\ref{app:weight_update}.

Each dynamic iteration performs one synchronized block update. The ball-level spectral embedding and the residual $L_p$ of each base kernel are first computed from the current weights. Eq.~\eqref{eq:weight_update} then updates the weights, after which $Z$ and the objective value are recomputed using the new $\sigma$. The granular-ball partition, kernel weights, spectral embedding, and objective value therefore describe the same state.

 The change in the objective is
\begin{equation}
	\Delta^{(t)}
	=
	|J^{(t)}-J^{(t-1)}|.
	\label{eq:outer_delta}
\end{equation}
The dynamic optimization terminates when $\Delta^{(t)}\leq\varepsilon$ or when the maximum number of iterations is reached. Convergence is evaluated before the next membership update, so the returned partition, weights, embedding, and objective value remain synchronized.

If the stopping condition is not met, the latest weights define the sample-level fused kernel
\begin{equation}
	K_{\sigma^{(t+1)}}
	=
	\sum_{p=1}^{P}
	(\sigma_p^{(t+1)})^2
	K^{(p)}.
	\label{eq:dynamic_fused_kernel}
\end{equation}
Using the $M$ centers computed from the current partition in this fused kernel space, all samples undergo one fixed-$M$ kernel $k$-means assignment update:
\begin{equation}
	B_l^{(t+1)}
	=
	\argmin_{1\leq b\leq M}
	\left\|
	\phi_{\sigma^{(t+1)}}(x_i)
	-
	C_b^{(t+1)}
	\right\|_{\mathcal H}^2,
	\quad x_i\in\mathcal X.
	\label{eq:dynamic_reassignment}
\end{equation}
%This step changes only sample memberships; it neither repeats center-consistency splitting nor changes $M$. If an empty ball appears, a sample with a large current assignment distance is reassigned to it so that every ball remains nonempty. The new memberships are then used to reconstruct $G$, $D$, and all $\widehat{K}_B^{(p)}$ before the next ball-level MKKM update. Alternating kernel-weight learning with fixed-count kernel-space reassignment allows the partition to track the changing fused kernel space.
This step updates only sample memberships while keeping $M$ fixed, without repeating center-consistency splitting. Empty balls are repaired by reassigning samples with large assignment distances. The updated memberships are then used to rebuild $G$, $D$, and all $\widehat{K}_{B^{(t+1)}}^{(p)}$ for the next ball-level MKKM update. Alternating kernel-weight learning with fixed-count reassignment enables the partition to track the evolving fused kernel space.

After convergence, the sample-level embedding is recovered from the final ball-level embedding as
\begin{equation}
	 H
	=
	G D^{-1/2} Z.
	\label{eq:sample_embedding}
\end{equation}
Each row of $H$ is $\ell_2$-normalized, and Euclidean $k$-means with $c$ clusters is applied to the normalized embedding. The complete procedure is listed in Algorithm~\ref{alg:dkgb_mkkm} in the Appendix.

The algorithm begins with uniform kernel weights and constructs the initial granular-ball partition using GBCT in the corresponding fused kernel space. It then fixes the number of balls and builds the sample-to-ball indicator, the ball-size matrix, and the weighted granular-ball kernel for each base kernel. During dynamic optimization, one synchronized spectral-and-weight update is followed, when necessary, by one kernel-space membership update at a fixed ball count. After convergence, the final ball-level embedding is lifted to the sample level and clustered. Thus, DK-GBMKKM couples ball-level multiple kernel learning with dynamic kernel-space reassignment while retaining a compact spectral problem. Detailed algorithmic procedures and time-complexity analysis are provided in the Appendix.

\section{Experimental Design and Results}
\label{sec:experiments}

We compare DK-GBMKKM with recent multiple kernel clustering baselines on 12 public datasets and examine convergence through changes in its objective value. All experiments were conducted in MATLAB R2025b.

\subsection{Setup}

% The experiments use 12 public datasets commonly adopted in feature-selection and clustering studies: arcene, Carcinom, lung\_discrete, ORL, orlraws10P, pixraw10P, SMK\_CAN\_187, TOX\_171, warpAR10P, warpPIE10P, COIL20, and Isolet. They cover gene expression, object images, speech, and face images~\cite{guyon2004result,samaria1994parameterisation,nene1996columbia,cole1990isolet,li2018feature,asuFeatureDatasets}. Several high-dimensional feature-selection datasets are the curated versions released by Feature Selection @ ASU~\cite{asuFeatureDatasets}. Ground-truth labels are used only to set the number of clusters and compute external evaluation metrics; they are not used for kernel construction, granular-ball generation, kernel-weight learning, or spectral embedding. Missing values are replaced with zero. Each feature is then z-score standardized, and every sample is $\ell_2$-normalized. Dataset details are reported in Table~\ref{tab:datasets_revised} in the Appendix.

\textbf{Datasets:} Experiments are conducted on 12 public datasets covering gene expression, object, speech, and face  data~\cite{guyon2004result,samaria1994orl,nene1996coil20,cole1990isolet,li2018feature,asuFeatureDatasets}, including several high-dimensional datasets from Feature Selection @ ASU~\cite{asuFeatureDatasets}. Ground-truth labels are used only to determine the number of clusters and compute evaluation metrics. Missing values are set to zero, followed by z-score standardization and sample-wise $\ell_2$ normalization. Dataset statistics are provided in Table~\ref{tab:datasets_revised} of the Appendix.

\textbf{Base kernels:}
We construct 12 base kernels: seven radial basis function (RBF) kernels, four polynomial kernels, and one cosine kernel. 
% The RBF kernels are
% \begin{equation}
	% K_{ij}^{(t)}
	% =
	% \exp\!\left(-\frac{\lVert x_i-x_j\rVert_2^2}{t\delta}\right),
	% \end{equation}
% where $t\in\{0.01,0.05,0.1,1,10,50,100\}$ and $\delta$ is the mean squared distance over all sample pairs. 
The RBF scale parameters are set to $t\in\{0.01,0.05,0.1,1,10,50,100\}$. 
The polynomial kernels are $(x_i^\top x_j+a)^b$, where $a\in\{0,1\}$ and $b\in\{2,4\}$. After row normalization, the cosine kernel is $x_i^\top x_j$. To ensure identical inputs across methods, every base kernel is cleaned of nonfinite entries, symmetrized, centered, and diagonal-normalized.

\textbf{Compared methods and parameter settings:}
We compare DK-GBMKKM with seven representative baselines: SMKC~\cite{liang2024smkc}, AASC~\cite{huang2012aasc}, MKKM~\cite{huang2012mkkm}, SimpleMKKM~\cite{liu2023simplemkkm}, RMKKM~\cite{du2015rmkkm}, MKKM-SR~\cite{lu2022mkkmsr}, and GB-MKKM~\cite{xia2025GBMKKM}. All methods use the same preprocessing, 12 base kernels, and number of clusters. Baseline parameters follow the recommended settings in the corresponding papers or public implementations.

\textbf{Evaluation metrics:}
Clustering quality is evaluated using clustering accuracy (ACC), normalized mutual information (NMI), Purity, and the adjusted Rand index (ARI). Higher values indicate better performance for all four metrics.

\subsection{Clustering Performance}

% Table~\ref{tab:average_results_revised} summarizes the average results over the 12 datasets. Figure~\ref{fig:acc_comparison} and the additional comparisons in Appendix~\ref{app:detailed_results} report the dataset-level results. DKGB-MKKM ranks first on all four average metrics, achieving ACC, NMI, Purity, and ARI values of 0.6717, 0.5929, 0.6848, and 0.4647, respectively. Its absolute gains over the second-best average are 0.0574, 0.0521, 0.0522, and 0.0590. Counting the best result on each dataset, DKGB-MKKM ranks first eight times for ACC and NMI and nine times for Purity and ARI. Its advantage is therefore not driven by extreme results on only a few datasets.

\begin{figure}[t]
	\centering
	\includegraphics[width=\columnwidth]{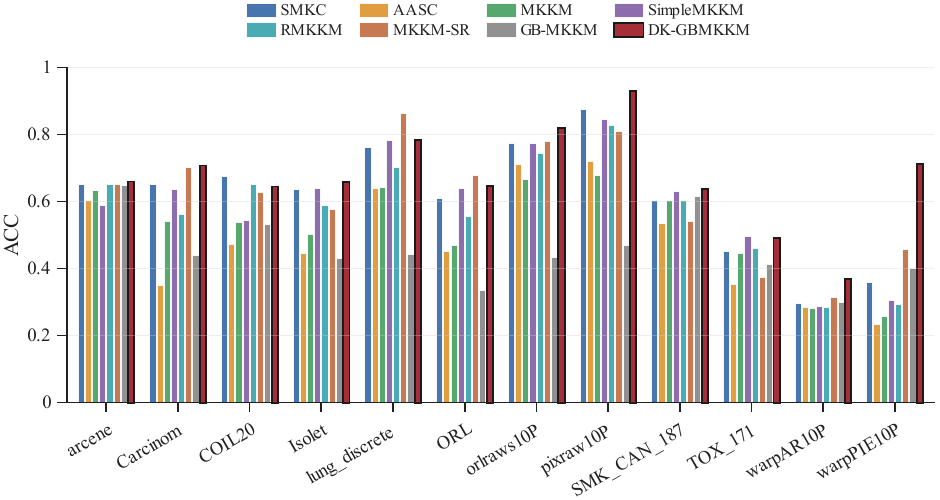}
	\caption{ACC of eight methods on 12 datasets.}
	\label{fig:acc_comparison}
\end{figure}

Table~\ref{tab:average_results_revised} summarizes the average results over 12 datasets, with detailed comparisons shown in Fig.~\ref{fig:acc_comparison} and Appendix~\ref{app:detailed_results}. DK-GBMKKM ranks first on all four average metrics, achieving 0.6717 ACC, 0.5929 NMI, 0.6848 Purity, and 0.4647 ARI, with gains of 0.0574, 0.0521, 0.0522, and 0.0590 over the second-best averages. It also records the most dataset-level wins, confirming that the improvement is consistent rather than dataset-specific.

\begin{table}[t]
	\centering
	\caption{Average clustering performance on 12 datasets.}
	\label{tab:average_results_revised}
\footnotesize  % ±È \small ¸üÐ¡
	\begin{tabular}{lcccc}
		\toprule
		Method & ACC & NMI & Purity & ARI\\
		\midrule
		SMKC & 0.6118 & 0.5363 & 0.6321 & 0.3965\\
		AASC & 0.4837 & 0.4200 & 0.5213 & 0.2561\\
		MKKM & 0.5212 & 0.4855 & 0.5560 & 0.3103\\
		SimpleMKKM & 0.5978 & 0.5165 & 0.6206 & 0.3743\\
		RMKKM & 0.5769 & 0.5154 & 0.6063 & 0.3595\\
		MKKM-SR & \underline{0.6143} & \underline{0.5408} & \underline{0.6327} & \underline{0.4058}\\
		GB-MKKM & 0.4555 & 0.3362 & 0.4806 & 0.1670\\
		DK-GBMKKM & \textbf{0.6717} & \textbf{0.5929} & \textbf{0.6848} & \textbf{0.4647}\\
		\bottomrule
	\end{tabular}
\end{table}

%\begin{table}[t]
%	\centering
%	\caption{Average clustering performance on 12 datasets.}
%	\label{tab:average_results_revised}
%	\resizebox{\columnwidth}{!}{%
%		\begin{tabular}{lcccc}
%			\toprule
%			Method & ACC & NMI & Purity & ARI\\
%			\midrule
%			SMKC & 0.6118 & 0.5363 & 0.6321 & 0.3965\\
%			AASC & 0.4837 & 0.4200 & 0.5213 & 0.2561\\
%			MKKM & 0.5212 & 0.4855 & 0.5560 & 0.3103\\
%			SimpleMKKM & 0.5978 & 0.5165 & 0.6206 & 0.3743\\
%			RMKKM & 0.5769 & 0.5154 & 0.6063 & 0.3595\\
%			MKKM-SR & \underline{0.6143} & \underline{0.5408} & \underline{0.6327} & \underline{0.4058}\\
%			GB-MKKM & 0.4555 & 0.3362 & 0.4806 & 0.1670\\
%			DKGB-MKKM & \textbf{0.6717} & \textbf{0.5929} & \textbf{0.6848} & \textbf{0.4647}\\
%			\bottomrule
%		\end{tabular}%
%	}
%\end{table}

% At the dataset level, DKGB-MKKM achieves the best value on all four metrics for orlraws10P, pixraw10P, warpAR10P, and warpPIE10P, indicating that dynamic kernel-space granular balls are well suited to data with complex image variations and high-dimensional representations. The improvement on warpPIE10P is particularly marked: compared with the runner-up MKKM-SR, DKGB-MKKM improves ACC, NMI, Purity, and ARI by 0.2558, 0.1916, 0.2028, and 0.2402, respectively. This result suggests that adapting ball boundaries as the fused-kernel geometry changes reduces the mismatch between a fixed input-space partition and the current kernel-space structure.
DK-GBMKKM achieves the best results on all four metrics for orlraws10P, pixraw10P, warpAR10P, and warpPIE10P. On warpPIE10P, it improves ACC, NMI, Purity, and ARI over the runner-up MKKM-SR by 0.2558, 0.1916, 0.2028, and 0.2402, respectively. This result suggests that adapting ball boundaries as the fused-kernel geometry changes reduces the mismatch between a fixed input-space partition and the current kernel-space structure.

DK-GBMKKM is not uniformly superior on every dataset. RMKKM obtains the highest NMI and Purity on COIL20, and MKKM-SR remains competitive on lung\_discrete and ORL. When the local structure in the input space is already stable or spectral rotation sufficiently aligns the embedding with discrete labels, dynamic ball updates may provide limited additional benefit. Nevertheless, the leading averages and consistent gains across most datasets support the effectiveness of DK-GBMKKM while also revealing its dependence on data geometry.

\subsection{Convergence Analysis}

\begin{figure}[t]
	\centering
	\includegraphics[width=\columnwidth]{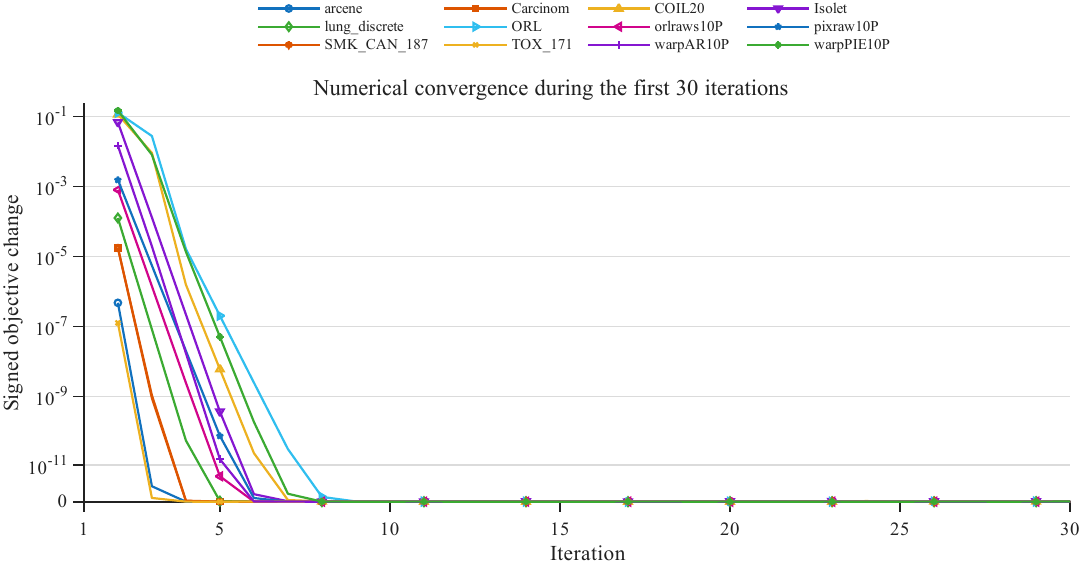}
	\caption{Signed change in the DK-GBMKKM objective over the first 30 iterations on 12 datasets.}
	\label{fig:objective_convergence}
\end{figure}

To examine whether dynamic membership updates induce persistent oscillations, we run DK-GBMKKM once on each dataset with early stopping disabled and record 30 consecutive objective values. 
% Define
% \begin{equation}
% \Delta J^{(t)}
% =
% J^{(t)}-J^{(t-1)}.
% \end{equation}
Fig.~\ref{fig:objective_convergence} plots the change $\Delta J^{(t)}$ over the first 30 iterations on all 12 datasets. A symmetric logarithmic scale displays both positive and negative changes, including fluctuations near machine precision.

All curves approach zero within the first few iterations. After iteration 20, $\Delta J^{(t)}$ remains below $5\times10^{-14}$ on every dataset, with no sustained oscillation above this scale. Under the current datasets and parameter settings, dynamic granular-ball reassignment therefore causes no observable late-stage oscillation of the objective.

\section{Conclusion}
\label{sec:conclusion}
This paper proposed DK-GBMKKM, which constructs and dynamically updates granular balls in the fused kernel space to reduce the mismatch between granular-ball representation and multiple kernel learning. By coupling ball-level MKKM optimization with kernel-space sample reassignment, the method enables adaptive partitioning under evolving kernel geometry. Experimental results verify its effectiveness. Future work will focus on low-rank kernel approximation and adaptive granularity for large-scale clustering.

%\section*{Acknowledgment}
%
%This work was supported in part by the National Natural Science Foundation of China under Grant 62221005, Grant 62450043, Grant 62222601, and Grant 62176033; in part by Chongqing University of Posts and Telecommunications Ph.D. Innovative Talent Project under Grant BYJS202403; and in part by Chongqing Natural Science Foundation Innovation and Development Joint Fund under Grant CSTB2025NSCQ-LZX0141.

\clearpage
\bibliographystyle{IEEEtran}
\bibliography{DKGB_MKKM_refs}

\clearpage
\appendices
\raggedbottom

\section{Experimental Dataset Details}

\begin{table}[H]
	\centering
	\caption{Information on the 12 experimental datasets.}
	\label{tab:datasets_revised}
	\small
	\setlength{\tabcolsep}{5pt}
	\begin{tabular}{lrrr}
		\toprule
		Dataset & Samples & Features & Classes\\
		\midrule
		arcene          & 200   & 10,000 & 2\\
		Carcinom        & 174   & 9,182  & 11\\
		COIL20          & 1,440 & 1,024  & 20\\
		Isolet          & 1,560 & 617    & 26\\
		lung\_discrete  & 73    & 325    & 7\\
		ORL             & 400   & 1,024  & 40\\
		orlraws10P      & 100   & 10,304 & 10\\
		pixraw10P       & 100   & 10,000 & 10\\
		SMK\_CAN\_187   & 187   & 19,993 & 2\\
		TOX\_171        & 171   & 5,748  & 4\\
		warpAR10P       & 130   & 2,400  & 10\\
		warpPIE10P      & 210   & 2,420  & 10\\
		\bottomrule
	\end{tabular}
\end{table}

\section{Theoretical Properties of Granular-Ball Representations in the Fused Kernel Space}
\label{app:fusion_ball}
\label{app:fused_ball_center}

This section establishes the relationship between a granular-ball center in the fused kernel space and its counterparts in the base-kernel spaces. From Eq.~\eqref{eq:sample_fusion},
\begin{equation}
	K_{\sigma}
	=
	\sum_{p=1}^{P}\sigma_p^2K^{(p)}.
\end{equation}
Let $\phi_p(x)$ be an implicit feature map for the $p$th base kernel. An equivalent feature map for the fused kernel is
\begin{equation}
	\phi_{\sigma}(x)
	=
	\left[
	\sigma_1\phi_1(x)^\top,\ldots,
	\sigma_P\phi_P(x)^\top
	\right]^\top.
	\label{eq:fused_feature_mapping}
\end{equation}
Indeed,
\begin{equation}
	\begin{aligned}
		\left\langle\phi_{\sigma}(x_i),\phi_{\sigma}(x_j)\right\rangle
		&=
		\sum_{p=1}^{P}\sigma_p^2
		\left\langle\phi_p(x_i),\phi_p(x_j)\right\rangle\\
		&=
		\sum_{p=1}^{P}\sigma_p^2K^{(p)}(x_i,x_j)\\
		&=
		K_{\sigma}(x_i,x_j),
	\end{aligned}
\end{equation}
which verifies Eq.~\eqref{eq:fused_feature_mapping}.

For a granular ball $B_b$ generated in the fused kernel space, let $\mathcal I_b$ be its sample index set and $n_b$ its size. From Eq.~\eqref{eq:kernel_space_ball_center}, its center is
\begin{equation}
	C_b
	=
	\frac{1}{n_b}
	\sum_{x_i\in\mathcal I_b}
	\phi_{\sigma}(x_i).
\end{equation}
Substituting Eq.~\eqref{eq:fused_feature_mapping} gives
\begin{equation}
	\begin{aligned}
		C_b
		&=
		\frac{1}{n_b}
		\sum_{x_i\in\mathcal I_b}
		\left[
		\sigma_1\phi_1(x_i)^\top,\ldots,
		\sigma_P\phi_P(x_i)^\top
		\right]^\top\\
		&=
		\left[
		\sigma_1C_b^{(1)\top},\ldots,
		\sigma_PC_b^{(P)\top}
		\right]^\top,
	\end{aligned}
	\label{eq:fused_ball_center_decomposition}
\end{equation}
where
\begin{equation}
	C_b^{(p)}
	=
	\frac{1}{n_b}
	\sum_{x_i\in\mathcal I_b}
	\phi_p(x_i)
\end{equation}
is the center of $B_b$ in the $p$th base-kernel space. Therefore, for any two balls $B_a$ and $B_b$,
\begin{equation}
	\begin{aligned}
		\left\langle C_a,C_b\right\rangle
		&=
		\left\langle
		\left[
		\sigma_1C_a^{(1)\top},\ldots,
		\sigma_PC_a^{(P)\top}
		\right]^\top,
		\right.\\
		&\quad\left.
		\left[
		\sigma_1C_b^{(1)\top},\ldots,
		\sigma_PC_b^{(P)\top}
		\right]^\top
		\right\rangle\\
		&=
		\sum_{p=1}^{P}\sigma_p^2
		\left\langle C_a^{(p)},C_b^{(p)}\right\rangle.
	\end{aligned}
	\label{eq:fused_ball_center_inner_product}
\end{equation}
Eq.~\eqref{eq:fused_ball_center_inner_product} shows that center relations in the fused kernel space are weighted combinations of the corresponding center relations under the same partition in the base-kernel spaces. A single partition can therefore be generated in the current fused kernel space and shared across all base kernels.

Since
\begin{equation}
	\left\langle C_a^{(p)},C_b^{(p)}\right\rangle
	=
	\overline K_B^{(p)}(a,b),
\end{equation}
we further have
\begin{equation}
	\left\langle C_a,C_b\right\rangle
	=
	\sum_{p=1}^{P}\sigma_p^2
	\overline K_B^{(p)}(a,b).
	\label{eq:fused_average_ball_kernel_element}
\end{equation}
For a fixed granular-ball partition, the similarities between fused-space ball centers are thus equivalent to the weighted combination of the corresponding base-space center similarities.

\section{Properties of the Weighted Granular-Ball Kernel}
\label{app:weighted_kernel}

\subsection{Proof of Proposition 1}
\label{app:prof1}

From Eq.~\eqref{eq:kernel_space_ball_center}, the inner product between the centers of balls $B_a$ and $B_b$ in the $p$th base-kernel space is
\begin{equation}
	\begin{aligned}
		\left\langle C_a^{(p)},C_b^{(p)}\right\rangle
		&=
		\left\langle
		\frac{1}{n_a}
		\sum_{x_i\in\mathcal I_a}\phi_p(x_i),
		\frac{1}{n_b}
		\sum_{x_j\in\mathcal I_b}\phi_p(x_j)
		\right\rangle\\
		&=
		\frac{1}{n_an_b}
		\sum_{x_i\in\mathcal I_a}
		\sum_{x_j\in\mathcal I_b}
		K^{(p)}(x_i,x_j).
	\end{aligned}
	\label{eq:appendix_ball_center_kernel}
\end{equation}
Collecting all pairwise center inner products gives
\begin{equation}
	\overline K_B^{(p)}
	=
	D^{-1}G^\top K^{(p)}GD^{-1}.
\end{equation}
Weighting this average kernel by the ball sizes yields
\begin{equation}
	\begin{aligned}
		\widehat K_B^{(p)}
		&=
		D^{1/2}\overline K_B^{(p)}D^{1/2}\\
		&=
		D^{-1/2}G^\top K^{(p)}GD^{-1/2}.
	\end{aligned}
\end{equation}
Consequently,
\begin{equation}
	\widehat K_B^{(p)}(a,b)
	=
	\sqrt{n_an_b}
	\left\langle C_a^{(p)},C_b^{(p)}\right\rangle.
\end{equation}

Let $Q=GD^{-1/2}$. Then
\begin{equation}
	\widehat K_B^{(p)}
	=
	Q^\top K^{(p)}Q.
\end{equation}
For any $\mathbf z\in\mathbb R^M$, if $K^{(p)}\succeq0$, then
\begin{equation}
	\begin{aligned}
		\mathbf z^\top\widehat K_B^{(p)}\mathbf z
		&=
		\mathbf z^\top Q^\top K^{(p)}Q\mathbf z\\
		&=
		(Q\mathbf z)^\top K^{(p)}(Q\mathbf z)
		\geq0.
	\end{aligned}
\end{equation}
Hence $\widehat K_B^{(p)}\succeq0$. Because $\sigma_p^2\geq0$,
\begin{equation}
	\widehat K_{B,\sigma}
	=
	\sum_{p=1}^{P}\sigma_p^2\widehat K_B^{(p)}
	\succeq0.
\end{equation}

\subsection{Proof of Proposition 2}
\label{app:prof2}

Using $H=QZ$, $Q^\top Q=I_M$, and $Z^\top Z=I_c$, we obtain
\begin{equation}
	\begin{aligned}
		H^\top H
		&=
		Z^\top Q^\top QZ
		=
		Z^\top Z
		=
		I_c.
	\end{aligned}
\end{equation}
For any base kernel $K^{(p)}$,
\begin{equation}
	\begin{aligned}
		\tr(H^\top K^{(p)}H)
		&=
		\tr\!\left(
		Z^\top Q^\top K^{(p)}QZ
		\right)\\
		&=
		\tr\!\left(
		Z^\top\widehat K_B^{(p)}Z
		\right),
	\end{aligned}
\end{equation}
which proves Eq.~\eqref{eq:alignment_equivalence}.

\section{Derivation of the Kernel-Weight Update}
\label{app:weight_update}

For a fixed ball-level spectral embedding $Z$, the kernel-weight subproblem in Eq.~\eqref{eq:ball_mkkm_objective} is
\begin{equation}
	\min_{\sigma}
	\sum_{p=1}^{P}\sigma_p^2L_p,
	\quad
	\mathrm{s.t.}\quad
	\sigma_p\geq0,
	\quad
	\sum_{p=1}^{P}\sigma_p=1,
\end{equation}
where
\begin{equation}
	L_p
	=
	\operatorname{Tr}(\widehat K_B^{(p)})
	-
	\operatorname{Tr}(Z^\top\widehat K_B^{(p)}Z)
\end{equation}
is the clustering residual associated with the $p$th base granular-ball kernel. Since $L_p\geq0$, the problem is a convex quadratic program over the probability simplex.

The Lagrangian is
\begin{equation}
	\mathcal L(\sigma,\lambda)
	=
	\sum_{p=1}^{P}\sigma_p^2L_p
	-
	\lambda\left(\sum_{p=1}^{P}\sigma_p-1\right),
\end{equation}
where $\lambda$ is the multiplier for the equality constraint. Setting the derivative with respect to $\sigma_p$ to zero gives
\begin{equation}
	\frac{\partial\mathcal L}{\partial\sigma_p}
	=
	2\sigma_pL_p-\lambda
	=
	0,
\end{equation}
and therefore
\begin{equation}
	\sigma_p
	=
	\frac{\lambda}{2L_p}.
\end{equation}
The normalization constraint yields
\begin{equation}
	\frac{\lambda}{2}
	\sum_{p=1}^{P}\frac{1}{L_p}
	=
	1,
\end{equation}
so that
\begin{equation}
	\frac{\lambda}{2}
	=
	\frac{1}{\sum_{q=1}^{P}1/L_q}.
\end{equation}
Substitution gives the optimal weight
\begin{equation}
	\sigma_p
	=
	\frac{1/L_p}{\sum_{q=1}^{P}1/L_q}.
\end{equation}
For numerical stability when a residual approaches zero, we use
\begin{equation}
	\widetilde L_p
	=
	\max(L_p,\epsilon),
\end{equation}
where $\epsilon$ is a small positive constant. The resulting update is
\begin{equation}
	\sigma_p^{+}
	=
	\frac{1/\widetilde L_p}
	{\sum_{q=1}^{P}1/\widetilde L_q},
\end{equation}
which is Eq.~\eqref{eq:weight_update}.

\section{DK-GBMKKM Optimization Procedure}
\label{app:algorithm}

\begin{algorithm}
	\caption{DK-GBMKKM}
	\label{alg:dkgb_mkkm}
	\begin{algorithmic}[1]
		\REQUIRE Base kernels $\{K^{(p)}\}_{p=1}^{P}$, number of clusters $c$, consistency coefficient $\eta$, minimum child-ball size $n_{\min}$, maximum number of ball-refinement rounds $T_{\mathrm{split}}$, tolerance $\varepsilon$, and maximum number of iterations $T$.
		\ENSURE Cluster labels $\mathbf y$, kernel weights $\sigma$, and granular-ball set $\mathcal B$.
		
		\STATE Initialize $\sigma_p^{(0)}=1/P$ and construct $K_{\sigma^{(0)}}$.
		\STATE Run GBCT on the initial fused kernel to obtain a complete granular-ball partition.
		\STATE Fix the resulting number of balls $M$ and construct $G$, $D$, and $\{\widehat K_B^{(p)}\}_{p=1}^{P}$.
		
		\FOR{$t=1,2,\ldots,T$}
		\STATE Construct $\widehat K_{B^{(t)},\sigma^{(t)}}$ from the current $\sigma^{(t)}$.
		\STATE Compute $Z^{(t)}=\operatorname{Spec}(\widehat K_{B^{(t)},\sigma^{(t)}},c)$ and the residual $L_p$ of each base kernel.
		\STATE Update the kernel weights with Eq.~\eqref{eq:weight_update} to obtain $\sigma^{(t+1)}$.
		\STATE Compute objective $J^{(t)}$.
		
		\IF{$t>1$ and $|J^{(t)}-J^{(t-1)}|\leq\varepsilon$}
		\STATE \textbf{break}
		\ENDIF
		
		\IF{$t<T$}
		\STATE Construct the latest sample-level fused kernel $K_{\sigma^{(t+1)}}$.
		\STATE Apply one fixed-$M$ kernel $k$-means assignment update to all samples using Eq.~\eqref{eq:dynamic_reassignment}, and repair any empty balls.
		\STATE Reconstruct $G$, $D$, and $\{\widehat K_{B^{t+1}}^{(p)}\}_{p=1}^{P}$ from the new memberships.
		\ENDIF
		\ENDFOR
		\STATE Recover $H=GD^{-1/2}Z$ using Eq.~\eqref{eq:sample_embedding}.
		\STATE Row-normalize $H$ in the $\ell_2$ norm and run Euclidean $k$-means with $c$ clusters to obtain $\mathbf y$.
		\RETURN $\mathbf y,\sigma,\mathcal B$.
	\end{algorithmic}
\end{algorithm}

\section{Time and Space Complexity}

Let $M$ be the final number of granular balls and $T$ the number of iterations. Given $P$ base kernels of size $n\times n$, constructing all weighted ball kernels requires $G^\top K^{(p)}G$. A dense implementation costs $O(Pn^2)$ time and $O(PM^2)$ memory for the ball kernels. Each synchronized MKKM update performs two eigendecompositions of an $M\times M$ matrix. A full decomposition costs $O(M^3)$, whereas iterative extraction of the leading $c$ eigenvectors typically costs approximately $O(M^2c)$. One fixed-count reassignment computes distances from $n$ samples to $M$ centers in the fused sample kernel and costs $O(n^2+nM)$ in a dense implementation.

Ignoring initialization constants, the dynamic phase has time complexity
$O\!\left(T(Pn^2+M^3+n^2+nM)\right)$; $M^3$ can be replaced by $M^2c$ when a partial eigensolver is used. If all base kernels remain in memory, the space complexity is $O(Pn^2+PM^2+nM)$. The current implementation therefore moves repeated spectral computations from matrices of order $n$ to matrices of order $M$. Because it retains full sample kernels and reaggregates them in each iteration, it is not a linear-memory method.

\section{Additional Dataset-Level Results}
\label{app:detailed_results}

\begin{figure}
	\centering
	\includegraphics[width=0.45\textwidth]{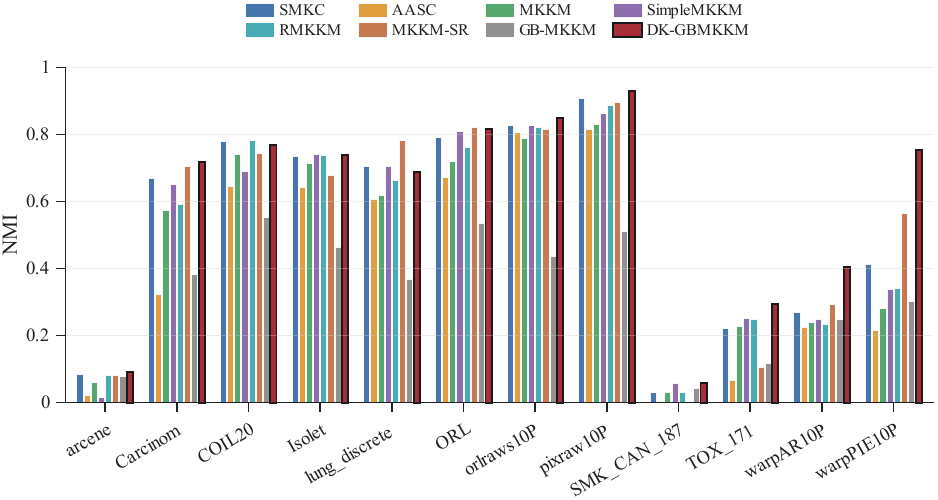}\\
	\includegraphics[width=0.45\textwidth]{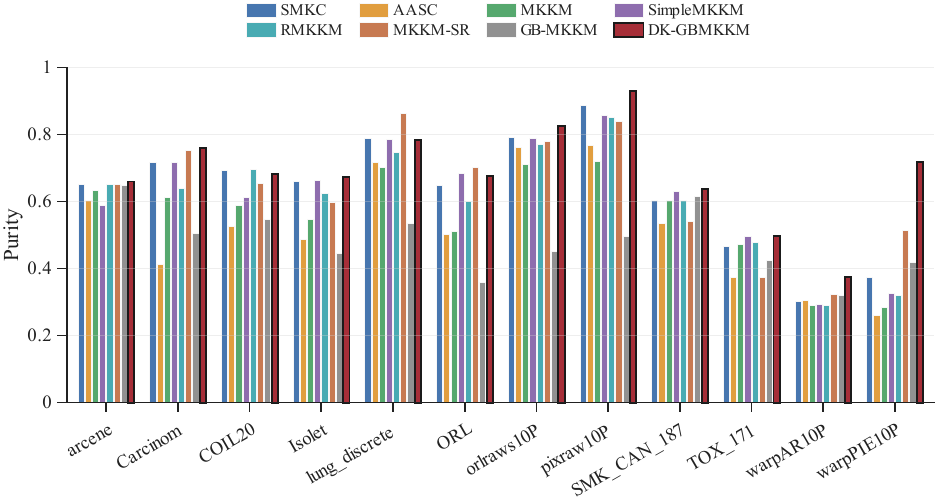}\\
	\includegraphics[width=0.5\textwidth]{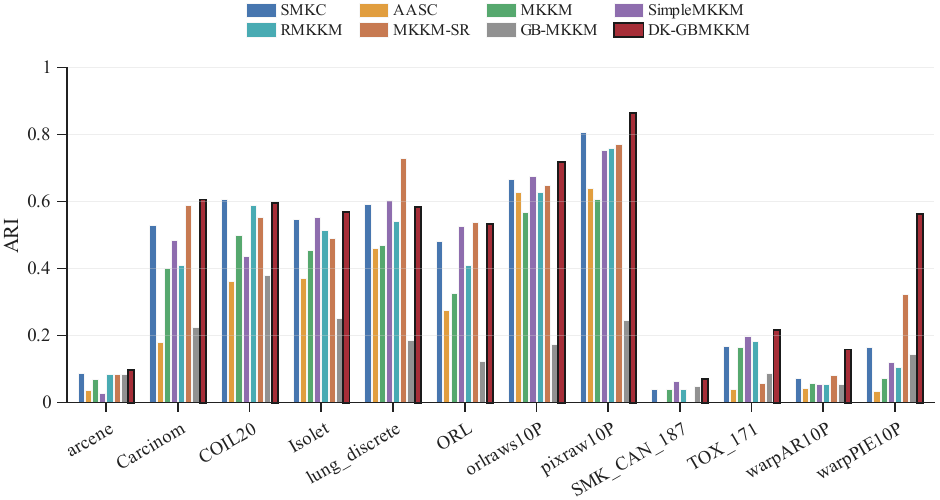}
	\caption{Comparison of eight methods on 12 datasets: NMI (top), Purity (middle), and ARI (bottom).}
	\label{fig:comparison}
\end{figure}

Fig.~\ref{fig:comparison} provide the dataset-level NMI, Purity, and ARI results that complement the ACC comparison in Fig.~\ref{fig:acc_comparison}.

% \begin{figure*}[t]
	% \centering
	% \includegraphics[width=0.96\textwidth]{nmi_comparison.pdf}
	% \caption{NMI of eight methods on 12 datasets.}
	% \label{fig:nmi_comparison}
	% \end{figure*}

% \begin{figure*}[t]
	% \centering
	% \includegraphics[width=0.96\textwidth]{purity_comparison.pdf}
	% \caption{Purity of eight methods on 12 datasets.}
	% \label{fig:purity_comparison}
	% \end{figure*}

% \begin{figure*}[t]
	% \centering
	% \includegraphics[width=0.96\textwidth]{ari_comparison.pdf}
	% \caption{ARI of eight methods on 12 datasets.}
	% \label{fig:ari_comparison}
	% \end{figure*}

%\FloatBarrier
\section{Statistical Tests and Significance Analysis}
\label{app:statistical_test}

To avoid drawing conclusions from mean values alone, we conduct nonparametric tests on the eight methods over the 12 datasets. Within each dataset, methods are ranked in descending order of metric value, with tied results assigned their average rank. A Friedman test first evaluates the null hypothesis that all eight methods have equal overall performance. Its statistic is
\begin{equation}
	\chi_F^2
	=
	\frac{12N}{k(k+1)}
	\left[
	\sum_{j=1}^{k}\bar R_j^2
	-
	\frac{k(k+1)^2}{4}
	\right],
\end{equation}
where $N=12$, $k=8$, and $\bar R_j$ is the average rank of method $j$. The Iman--Davenport statistic
\begin{equation}
	F_F
	=
	\frac{(N-1)\chi_F^2}{N(k-1)-\chi_F^2}fig:combined_cd
\end{equation}
provides a finite-sample correction. If the global null hypothesis is rejected, a Nemenyi post hoc test is applied. At significance level $\alpha=0.05$, the critical difference is
\begin{equation}
	\mathrm{CD}
	=
	q_{0.05}
	\sqrt{\frac{k(k+1)}{6N}}
	=
	3.031.
\end{equation}

As shown in Table~\ref{tab:statistical_tests}, the Iman--Davenport tests produce very small $p$-values for all four metrics: $3.71\times10^{-13}$ for ACC, $6.27\times10^{-14}$ for NMI, $7.20\times10^{-11}$ for Purity, and $1.10\times10^{-13}$ for ARI. All are well below 0.05, so the null hypothesis of equal performance is rejected for every metric. DK-GBMKKM also obtains the best average rank among the eight methods: 1.42 for ACC, 1.67 for NMI, 1.58 for Purity, and 1.42 for ARI. Its advantage in Table~\ref{tab:average_results_revised} is therefore consistent across datasets and is not an artifact of individual datasets or metric scales.

\begin{table}[H]
	\centering
	\caption{Friedman and Iman--Davenport test results.}
	\label{tab:statistical_tests}
	\small
	\begin{tabular}{lccc}
		\toprule
		Metric & $\chi_F^2$ & $F_F$ & $p$-value\\
		\midrule
		ACC    & 50.5347 & 16.6107 & $3.71\times10^{-13}$\\
		NMI    & 52.1042 & 17.9693 & $6.27\times10^{-14}$\\
		Purity & 45.3681 & 12.9180 & $7.20\times10^{-11}$\\
		ARI    & 51.6181 & 17.5344 & $1.10\times10^{-13}$\\
		\bottomrule
	\end{tabular}
\end{table}

\begin{figure*}[t]
	\centering
	\begin{subfigure}[b]{0.45\textwidth}
		\centering
		\includegraphics[width=\textwidth]{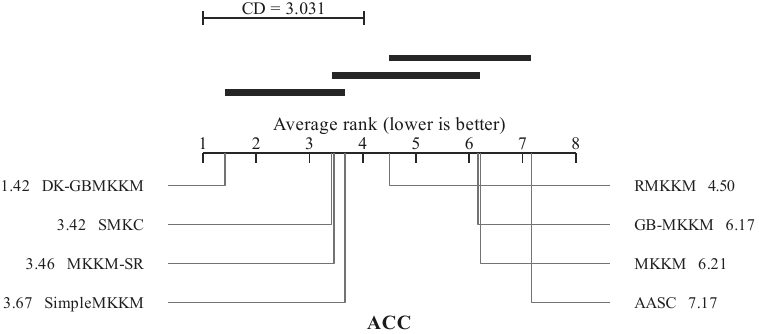}
		\caption{ACC}
		\label{fig:acc_cd}
	\end{subfigure}
	\hfill
	\begin{subfigure}[b]{0.45\textwidth}
		\centering
		\includegraphics[width=\textwidth]{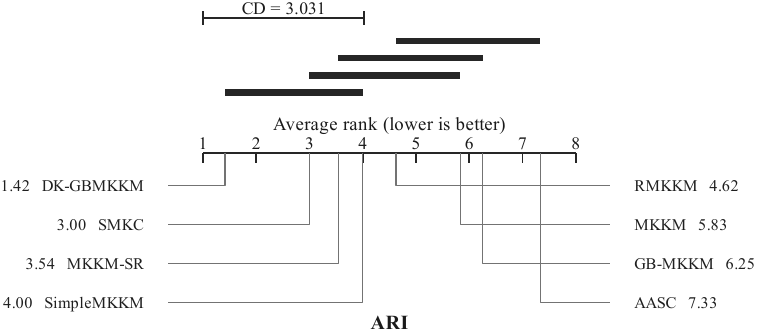}
		\caption{ARI}
		\label{fig:ari_cd}
	\end{subfigure}
	\par\medskip
	\begin{subfigure}[b]{0.45\textwidth}
		\centering
		\includegraphics[width=\textwidth]{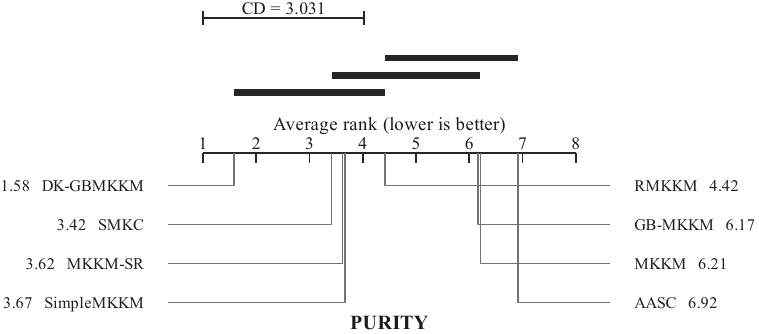}
		\caption{Purity}
		\label{fig:purity_cd}
	\end{subfigure}
	\hfill
	\begin{subfigure}[b]{0.45\textwidth}
		\centering
		\includegraphics[width=\textwidth]{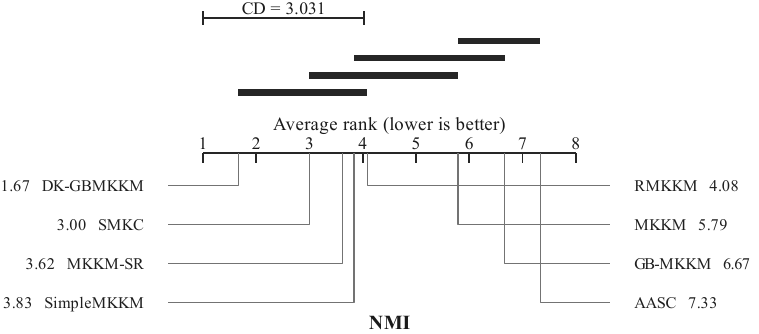}
		\caption{NMI}
		\label{fig:nmi_cd}
	\end{subfigure}
	\caption{Nemenyi critical-difference diagrams for four metrics. Lower average ranks indicate better performance; thick lines connect algorithms whose differences are not significant.}
	\label{fig:combined_cd}
\end{figure*}

Fig.~\ref{fig:combined_cd} shows the critical-difference diagrams for ACC, ARI, Purity, and NMI. All four use $\mathrm{CD}=3.031$ for 12 datasets and eight methods. DK-GBMKKM attains the lowest average rank for every metric, confirming its best overall performance across evaluation criteria.

The rank gaps between DK-GBMKKM and MKKM, GB-MKKM, and AASC exceed the critical difference for all four metrics, indicating a significant advantage over these conventional multiple kernel and granular-ball multiple kernel clustering methods. Relative to RMKKM, the difference is significant only for ACC. The gaps from SMKC, MKKM-SR, and SimpleMKKM are not significant for most metrics, showing that these strong baselines remain competitive. Nevertheless, DK-GBMKKM ranks first on all four metrics, supporting the effectiveness of dynamic granular-ball adjustment and kernel-space structural modeling.

\end{document}